\documentclass[letterpaper, 10 pt, conference]{ieeeconf}  

\IEEEoverridecommandlockouts                              

\title{\LARGE \bf
Sliding Touch-based Exploration \\ for Modeling Unknown Object Shape with Multi-fingered Hands

}

\author{Yiting Chen$^{1}$, Ahmet Ercan Tekden$^{1}$, Marc Peter Deisenroth$^{2}$, Yasemin Bekiroglu$^{1,2}$
\thanks{$^{1}$Department of Electrical Engineering, Chalmers University of Technology, Sweden. $^{2}$Department of Computer Science, University College London, U.K. This work was supported by Chalmers AI Research Center (CHAIR) and Chalmers Gender Initiative for Excellence (Genie). Email: \tt\small yitingch@student.chalmers.se}}
 \usepackage{threeparttable}
\usepackage{graphicx} 
\usepackage{float} 
\usepackage{subfigure} 
\usepackage{amsmath,amssymb,amsfonts}
\usepackage{cite}
\usepackage{amsmath}
\usepackage{mathrsfs}
\usepackage{algorithm}
\usepackage{color,soul}
\usepackage{xcolor}
\usepackage{algorithmic}

\begin{document}

\maketitle
\thispagestyle{empty}
\pagestyle{empty}

\begin{abstract}
Efficient and accurate 3D object shape reconstruction contributes significantly to the success of a robot's physical interaction with its environment. Acquiring accurate shape information about unknown objects is challenging, especially in unstructured environments, e.g. the vision sensors may only be able to provide a partial view. To address this issue, tactile sensors could be employed to extract local surface information for more robust unknown object shape estimation. In this paper, we propose a novel approach for efficient unknown 3D object shape exploration and reconstruction using a multi-fingered hand equipped with tactile sensors and a depth camera only providing a partial view. 
We present a multi-finger sliding touch strategy for efficient shape exploration using a Bayesian Optimization approach and a single-leader-multi-follower strategy for multi-finger smooth local surface perception. We evaluate our proposed method by estimating the 3D shape of objects from the YCB and OCRTOC datasets based on simulation and real robot experiments. 
The proposed approach yields successful reconstruction results relying on only a few continuous sliding touches. Experimental results demonstrate that our method is able to model unknown objects in an efficient and accurate way.

\end{abstract}

\section{INTRODUCTION}

Robotic physical interaction tasks, such as 
grasping and dexterous manipulation, benefit from knowing the 3D shape of the object. Though visual-only (RGB and depth) 3D object perception methods have been well studied during past years, robotic visual perception accuracy is reduced in presence of noisy, incomplete, or occluded data. Approaches rely on sufficient prior knowledge from pre-trained neural network models, which often fail due to uncertainties regarding unknown objects and unstructured environments \cite{fu2021single}. 

Humans make use of both visual and tactile information to perceive features of unknown objects 
\cite{helbig2007optimal}. Even when visual perception is heavily confined, humans can still explore target objects' size, shape, and texture using 
tactile sensing. Thanks to the advances in robotic tactile sensors, nowadays robots are enabled to have human-like tactile perception \cite{dahiya2009tactile, yang2022biomimetic}. Optical tactile sensors \cite{yuan2017gelsight, lambeta2020digit} are popular given their high-frequency, high resolution, and ability to capture fine details about the local contact. The development of tactile simulators \cite{wang2022tacto} has also contributed greatly to including tactile sensing in robotic applications. 
\begin{figure} [t]
    \centering
    \includegraphics[width=\linewidth]{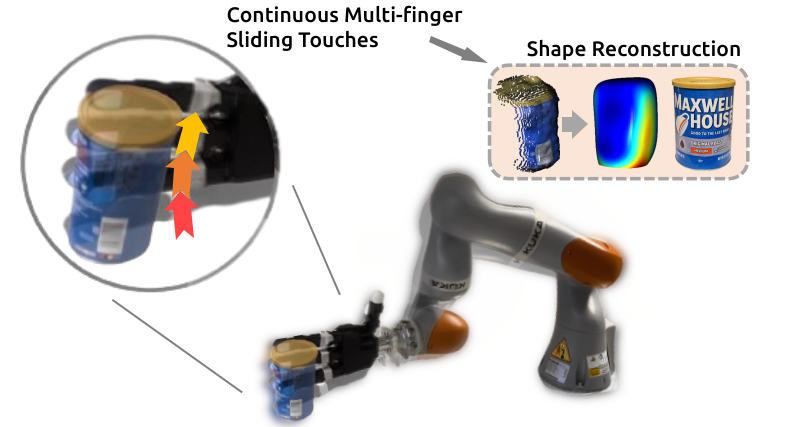}
    \caption{We propose a visuo-tactile perception approach based on sliding touches using a multi-fingered hand for unknown object modeling. A few sliding touches are illustrated in the upper left corner, using a multi-fingered hand visualized in gray for the first touch and black for the final touch. The hand is continuously moved in between these touches following the surface normal, reducing the uncertainty and increasing the accuracy of the initial object model that is fit based on visual data only. The initial incomplete point cloud obtained from the vision sensor, the final complete reconstructed object model (where blue color corresponds to low uncertainty), and the ground truth model respectively are seen in the upper right corner.}
    \label{system_setup}
\end{figure}
However, how to efficiently fuse both tactile and visual information for reconstructing an unknown object's shape sufficiently well for subsequent manipulation planning still remains an open problem. 
Tactile perception has been complementary to visual perception in most multi-modal settings. Existing methods represent tactile feedback as discrete points or a few pixels \cite{bjorkman2013enhancing, gandler, ilonen2014three, mahler2015gp, de2021simultaneous} during manipulation tasks. \cite{suresh2022shapemap, wang20183d} propose to use tactile images to predict local shape information, which further contributes to global shape reconstruction. However, all of the above methods require frequent re-planning of the robotic arm motion and are time-consuming in the exploration process because the robot needs to constantly re-establish new contacts with objects. 

Compared with discretely sampled interactions, tactile servoing aims at continuously applying a certain pressure on target objects to perform interaction tasks \cite{li2020review}, which allows robots to extract the geometric features smoothly. Tactile servoing includes important tasks, such as sliding a fingertip across an object’s surface \cite{li2013control}, which brings inspiration to robotic surface exploration. Previous works \cite{lepora2017exploratory, li2015visuo, driess2017active} have demonstrated that the robot can robustly follow the target object surface with one tactile sensor and further provides shape information such as the object's contour. Though tactile servoing with multi-fingered hands has been studied in robotic grasping \cite{liu2022multi}, the aforementioned tactile servoing methods for object surface exploration mainly only use a single tactile sensor.

One of the remaining key challenges is to fully harness the potential of the multi-fingered robotic hand to achieve human-like perception.  
Moreover, under the limited coverage of tactile sensors, making full use of the dexterity of each robotic fingertip for sensing the target object's surface will contribute significantly to perception efficiency. To tackle the above issues, we propose a single-leader-multi-follower sliding touch strategy for multi-fingered hands, which takes every fingertip into account for surface exploration.

\begin{figure}[t]
    \centering
    \includegraphics[width=0.8\linewidth]{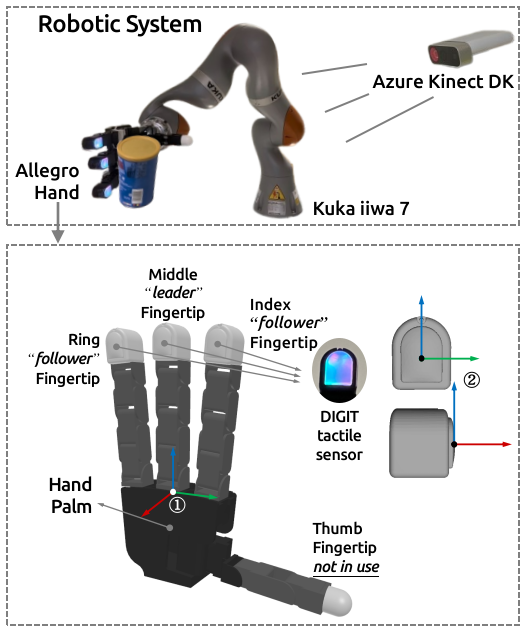}
    \caption{Robot setup, composed of a KUKA IIWA arm, Allegro Hand equipped with Digit tactile sensors, and an Azure Kinect scene camera. Frame 1 is rigidly attached to the hand palm to represent the palm pose and Frame 2 shows how local frames are rigidly attached to each fingertip.}
    \label{allegrohand}
\end{figure}

This paper presents a novel framework for reconstructing the 3D shape of unknown objects through visuo-tactile perception using a multi-fingered robotic hand.  First, we leverage the power of domain randomization from the tactile sensor simulator to learn a direct mapping between the tactile images from the DIGIT tactile sensors to the corresponding height maps. Then we propose a single-leader–multi-follower strategy for efficient tactile perception based on tactile servoing. The ``leader" fingertip provides tactile features, guiding the robotic hand during the sliding servoing movement. Meanwhile, the ``follower" fingertips expand the contact area by employing a fingertip adaptation strategy. We model the target object using a probabilistic representation, Gaussian process implicit surfaces (GPIS) \cite{eGPIS, khadivar2023online}, which can be further combined with query point guidance from Bayesian optimization (BOpt) \cite{driess2017active, de2021simultaneous} for sliding touch exploration. We show how combined visual perception and multi-finger tactile sensing improve the efficiency of shape exploration (shown in Fig. \ref{system_setup}) by significantly reducing the robotic arm motions. 

To the best of our knowledge, this is the first study that considers the continuous motion of a multi-fingered hand in visuo-tactile perception for the purpose of unknown object modeling through exploration. We present both simulated and real experiments, validating our proposed method of modeling unknown objects' global shapes with continuous robotic arm motion minimizing the required actions to reconstruct object models similar to the ground truth data.

Our main contributions can be summarized as follows: $\textbf{i)}$ We propose an efficient visuo-tactile perception approach based on continuous sliding touches with multi-fingered hands. Our approach provides an efficient way for surface exploration and significantly reduces the robotic arm motion and the required execution time. $\textbf{ii)}$ We present a hand-agnostic single-leader-multi-follower hand control strategy to combine tactile servoing and fingertip adaptation for smooth tactile sensing. This strategy fully exploits multi-fingered hand dexterity for maximizing the local contact area being perceived. $\textbf{iii)}$ We demonstrate that the visuo-tactile perception for unknown object modeling can be performed with limited arm motion in a smooth manner without re-establishing contact.

\section{System Setup and Problem Formulation \label{formulation}}
\subsection{System Setup}
\begin{figure} [htb]
    \centering
    \includegraphics[width=\linewidth]{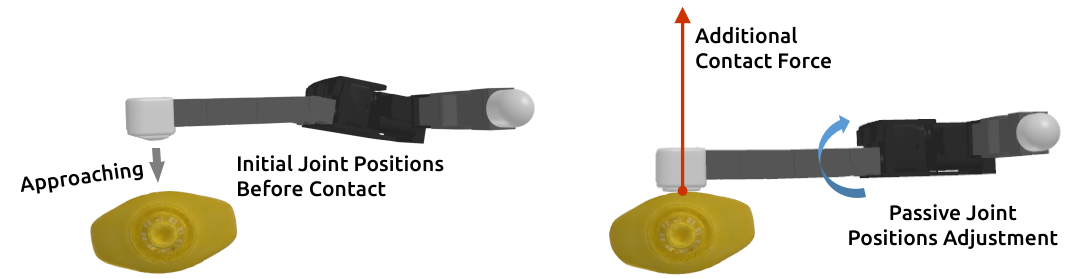}
    \caption{The initial hand configuration provides passive joint space for additional contact force between fingertips and the target object. During the sliding process, the hand configuration remains around zero-position to apply a constant contact force.}
    \label{allegro_contact}
\end{figure}

Fig. \ref{allegrohand} shows the multi-fingered hand used in our experiments to evaluate the proposed sliding touch approach. Each of the index, middle, and ring fingertips of the Allegro hand is equipped with DIGIT tactile sensor. We define the middle fingertip as the ``leader" and the index and ring fingertip as the ``follower", a detailed strategy will be introduced in Section \ref{sec_5}. The thumb fingertip is not in use for tactile sensing because we do not construct any force-closure configurations during exploration and only explore from the side of the object that is not visible to the scene camera. An impedance controller \cite{li2014learning} is deployed for finger control, which provides passive position-force adjustment during the interaction. Fingers are independent during contact and do not affect each other. The initial finger configuration is shown in Fig. \ref{allegro_contact}, a small joint position offset is set to the second joint of each finger.  After getting in contact with the target object, the fingers will passively move near the zero position which causes an additional force to press on the target object.

\subsection{Problem Formulation \label{formualtion}}
We address the problem of how to use a partial view of the depth camera and the sequential robotic tactile sliding touches to model the shape of a given unknown object. We consider that the underlying true shape can be represented using an implicit ``black-box" scalar function $\boldmath{f}$, which defines a manifold
\begin{equation}
    \mathcal{S} := \{x\in \mathbb{R}^3|f(x)=0 \}.
\end{equation}
We observe $f(x)$ from visual and tactile perception. As for the visual perception, we use the depth image $\textbf{D}\in \mathbb{R}^{H_1 \times W_1}$ from the partial view of the target object and the camera intrinsic and extrinsic matrix, by transferring $\textbf{D}$ into point cloud $\mathbf{P^w_d} = \{x^d_1, x^d_2, ..., x^d_{k} \}\in \mathbb{R}^3, k\leq H_1 \cdot W_1$ w.r.t the world frame after filtering with object segmentation mask, $f(x)$ can be observed as
\begin{equation}
        f(x^d_i)=0, \quad x^d_i \in \mathbf{P^w_d}.
\end{equation}

The unseen part of the object is perceived based on touch sensing. Given the tactile image $\mathbf{I} \in \mathbb{R}^{H_2\times W_2 \times 3}$ from the tactile sensor, we can estimate the height map of local contact $\mathbf{I} \rightarrow \mathbf{M} \in \mathbb{R}^{H_2\times W_2}$. Based on robotic forward kinematics and sensor intrinsic matrix, the height map is transferred from the fingertip's image frame to the tactile point cloud $\mathbf{M}\rightarrow \{x_1^w, x_2^w, ..., x_{n}^w\}, n \leq H_2\cdot W_2$ w.r.t the world frame after filtering with the binary contact area. 
We represent the motion of one sliding touch $\mathbf{T} = \{I_1, I_2, I_3, ..., I_m\} \rightarrow \{M_1, M_2, M_3, ..., M_m\}$ as a series of consecutive touch frames. The tactile point cloud $\mathbf{P^w_t}$ from one complete sliding touch is obtained as 
\begin{equation}
\begin{aligned}
      \mathbf{P^w_t} = \{x^t_1, x^t_2, ..., x^t_{k}\}\in \mathbb{R}^3, k\leq m\cdot H_2\cdot W_2.
\label{st}
\end{aligned}
\end{equation}
The value of $f(x)$ is given by
\begin{equation}
        f(x^t_j)=0, \quad x^t_j\in \mathbf{P^w_t}.
\end{equation}

Based on the observations $f(x)$, we use Gaussian Process Regression (GPR) to build an implicit surface (GPIS - Gaussian Process Implicit Surface) representation to estimate our target function $f$. The GPIS also provides a probabilistic description to further combine with BOpt to guide our robotic fingertip tactile perception. The sliding touch strategy will output sequential tactile point clouds $\mathbf{P^w_t}$ after each BOpt iteration for continuous $f(x)$ observation and leads to GPIS with lower uncertainty and more thorough modeling.

The rest of the paper is organized as follows: Section \ref{sec_4} presents the fingertip perception method for contact shape estimation and tactile feature extraction. Section 
\ref{sec_5} introduces the sliding touch strategy for efficient and smooth visuo-tactile perception. Section \ref{sec_6} presents experiments using sliding touches with simulation and a real robot, followed by conclusions in Section \ref{sec_7}.

\section{Fingertip Tactile Perception \label{sec_4}}

\subsection{Local Shape Estimation from Tactile Image}
The vision-based tactile sensor mainly consists of an embedded camera, an inner light source, and its surface gel \cite{lambeta2020digit, yuan2017gelsight}. The output tactile images reflect the surface deformation of the gel pad under special light conditions. We consider the task of recovering the height map $\textbf{M}\in \mathbb{R}^{H_2\times W_2}$ from the gel image $\textbf{I} \in \mathbb{R}^{H_2 \times W_2 \times3}$ as a similar problem of monocular depth prediction. 

\begin{figure}[htb]
    \centering
    \includegraphics[width=\linewidth]{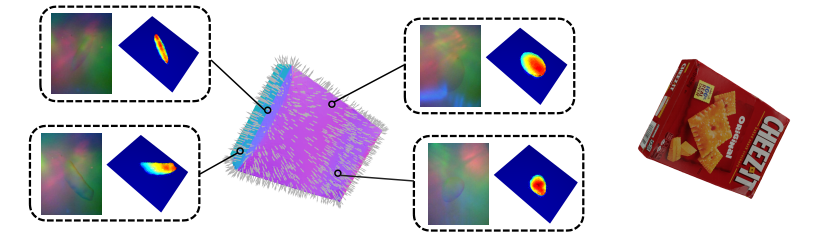}
    \caption{Four rendered tactile image-height map pairs with different calibrated backgrounds from real-world DIGIT sensors. We randomly select a surface normal as our reference direction and sample the contact surface with angle noise and randomized depth. In total, we generate {\color{black}{100,000 pairs}} of labeled data from 25 different objects for depth estimator training.}
    \label{sample_normals}
\end{figure}

\textbf{Data Collection in Simulation}: We learn the depth estimator in a supervised fashion with paired tactile images and their corresponding height maps. Since it is challenging to collect ground truth height maps in the real world, we collect the training data with TACTO \cite{wang2022tacto}, a flexible tactile sensor simulator with background calibration. We first calibrate the tactile simulator with different sensor backgrounds and uniformly sample tactile images across the object's surface. To simulate the real tactile perception situation, we sample the contact with different depth values $X\sim U[-0.2\,\text{mm}, -1.2\,\text{mm}]$ along surface normals with an orientation noise angle $\theta \sim \mathcal{N}(0, 30^{\circ})$. Rendered tactile images with different calibrated sensor backgrounds and corresponding height maps are shown in Fig. \ref{sample_normals}. We sample 4000 labeled tactile images on each of the 25 objects' mesh from the YCB dataset. 

\begin{figure} [htb]
    \centering
    \includegraphics[width=\linewidth]{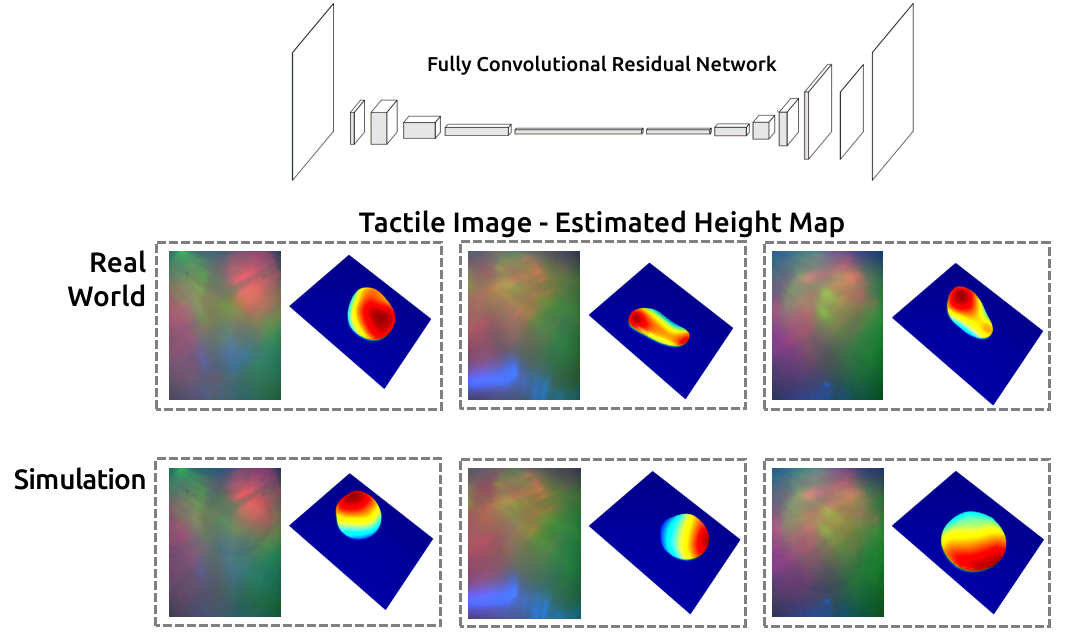}
    \caption{A fully convolutional residual network is adopted as our depth estimator. The upper line image pairs show the height map prediction results on real-world tactile images from different DIGIT sensors, and the bottom line image pairs show prediction results from simulated tactile images.}
    \label{real_world_exp}
\end{figure}

\textbf{Network Evaluation}: We adapt a modified fully convolutional residual network (FCRN) \cite{laina2016deeper} as our depth estimator, which uses a ResNet-50 as the backbone network and up-sampling blocks as the regression head. The depth estimator is responsible for $\textbf{I}\rightarrow \textbf{M}$ mapping. After 15 epochs of training with a learning rate of $2\text{e}^{-5}$, the prediction result in the real world and the network structure is shown in Fig. \ref{real_world_exp}.

\subsection{Tactile Feature Representation and Extraction}

\begin{figure}[htb]
    \centering
    \includegraphics[width=\linewidth]{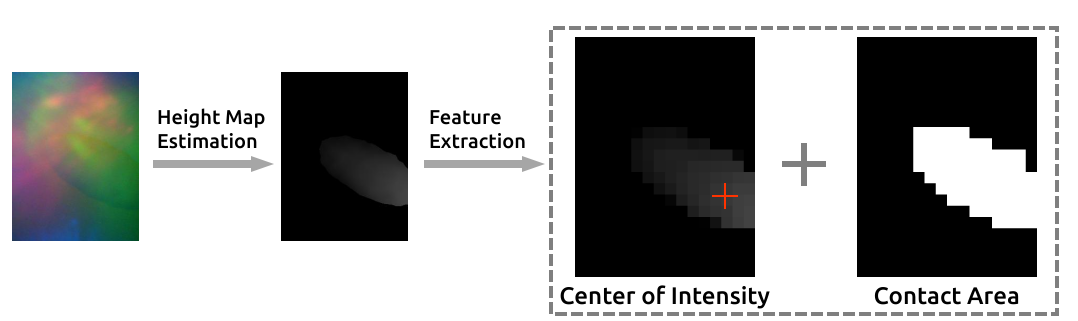}
    \caption{After estimating the height map from the tactile image, we extract the center of tactile intensity and contact area as our tactile feature for further servoing. The corresponding height map is resized into $32 \times 24$ for better visualization. The red cross in the left image from the dotted box denotes the extracted center of intensity and the right image from the dotted box demonstrates the binary contact area.}
    \label{tactile_feature}
\end{figure}

We extract the tactile feature $t_f$ from the estimated height map. Consider a height map $\textbf{M} \in \mathbb{R}^{h \times w}$ as demonstrated in Fig. \ref{tactile_feature}, we aim at extracting the center of intensity and the size of contact area as our tactile feature. 
The center of tactile intensity $(\overline{x}, \overline{y})$ is related to the contact pressure and the shape of the contact area, which can be obtained as 
\begin{equation}
    (\overline{x}, \overline{y}) = \frac{\sum^{h}_{i=1}\sum^{w}_{j=1}(x_i, y_j)m_{i,j}}{\sum^{h}_{i=1}\sum^{w}_{j=1}m_{i,j}}
\end{equation}
in which $x$ and $y$ denote the pixel index for columns and rows. The size of the contact area $c$ is computed by the total number of pixels with a value larger than $10^{-4}$. Finally, the tactile feature vector is concatenated in 3 dimensions as follows:
\begin{equation}
    t_f = \begin{bmatrix}
        \overline{x} \\
        \overline{y} \\
        c
    \end{bmatrix} = \begin{bmatrix}
        \text{Center of contact intensity along X} \\
        \text{Center of contact intensity along Y} \\
        \text{Contact area}
    \end{bmatrix}
\end{equation}
The tactile features extracted from fingers with different roles will be used to perform different tasks as follows:
\begin{itemize}
    \item \textbf{From the ``leader" fingertip}: Tactile feature will be used for hand palm twists and sliding direction calculation.
    \item \textbf{From the ``follower" fingertips}: Tactile feature will be used to acquire a larger contact area for tactile sensing.
\end{itemize}
The detailed mapping will be introduced in Section \ref{sliding_ada}.

\begin{figure}
    \centering
    \includegraphics[width=\linewidth]{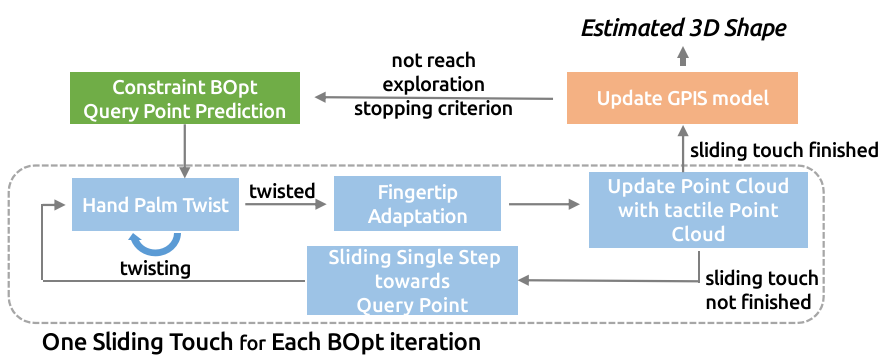}
    \caption{The online perception pipeline of our multi-fingered hand sliding touch strategy. Given the query point from each constrained BOpt iteration, first, the hand palm will twist based on the tactile feature from the ``leader" fingertip, and the ``follower" fingertips will adapt their pose given the current hand pose for a larger sensing area. After the visuo-tactile point cloud is updated, the hand will slide under the direction of obtained query point with a pre-defined step size.}
    \label{pipeline}
\end{figure}
\section{Sliding Touch for Shape Reconstruction\label{sec_5}}
\begin{figure*}[htbp]
\includegraphics[width=\textwidth]{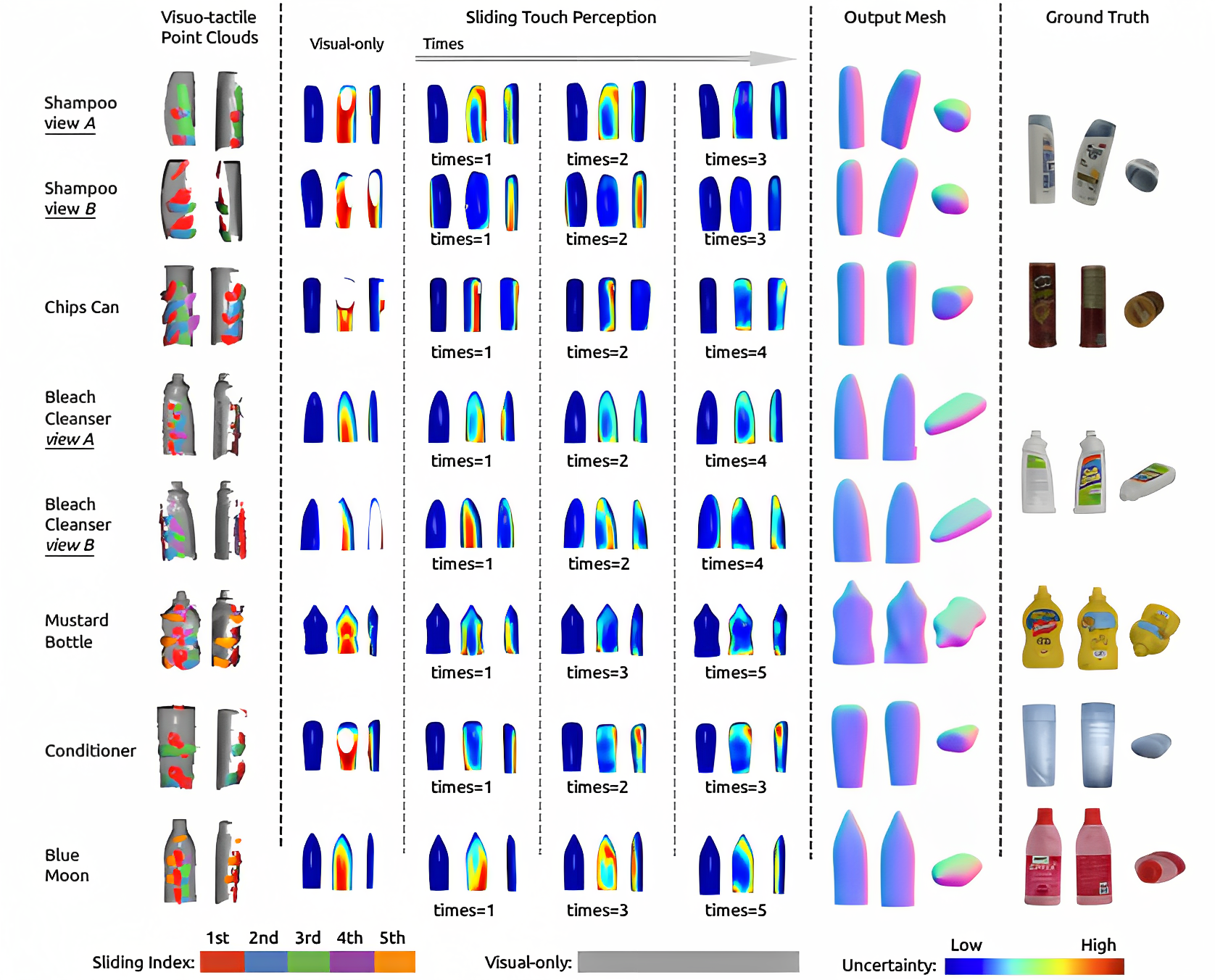}
\centering
\caption{The shape exploration and reconstruction process of 6 objects in the simulation. From left to right respectively: (I) the final combined visuo-tactile point cloud, where grey colored points represent the visual point cloud, and tactile point cloud generated by each sliding touch are colored with corresponding color map, (II) how the uncertainty map evolves during the sliding process (III) the final reconstructions and (IV) the ground truth models.}
\label{sim_results}
\end{figure*}

The pipeline of our proposed framework and the simulation results are illustrated in Fig. \ref{pipeline} and Fig. \ref{sim_results} respectively.
By fusing $P_{ti}^w$ with observed $P^w = \{P_d^w, P_{t1}^w, P_{t2}^w, ..., P_{t(i-1)}^w\} $ after each sliding touch, we are able to observe the target object's implicit function as introduced in Section \ref{formulation} and update our GPIS. The complete exploration process consists of several sliding touches $\{T_1, T_2, ..., T_k\}$, and one sliding touch consists of dozens of touch frames as previously defined in (\ref{st}).

\subsection{GPIS Shape Representation}
Based on the values $\boldsymbol{Y}$ observed from points $\boldsymbol{X} = P^w$, our goal is to reconstruct an implicit surface approximation of the object's shape while also providing uncertainty information for exploration guidance. The value of our GPIS $g^{\text{IS}}(x)$ is defined as follows:
\begin{equation}
  g^{\text{IS}}(x) \left\{
\begin{aligned}
 < & \quad 0 & \text{if $x$ is \textbf{below} the surface}\\
= & \quad 0 & \text{if $x$ is \textbf{on} the surface} \\
> & \quad 0 & \text{if $x$ is \textbf{above} the surface}
\end{aligned}
\right.
\label{gp_setting}
\end{equation}
which represents a mapping from point cloud to a scalar value $\mathbb{R}^3 \rightarrow \mathbb{R}$. We adopt the RBF\footnote[1]{For rectangular objects the kernel can be replaced by the thin-plate kernel, which leads to better performance.}
 (squared-exponential) kernel $k_{\text{RBF}}(\mathbf{x_1}, \mathbf{x_2}) = \exp \left( -\frac{1}{2} (\mathbf{x_1} - \mathbf{x_2})^\top \Theta^{-2} (\mathbf{x_1} - \mathbf{x_2}) \right)$ to approximate the target implicit function $f$, which leads to good reconstruction results for the experiment objects. Given a query point $x_{*} \in \mathbb{R}^3$, the $g^{\text{IS}}$ computes its posterior mean $\overline{g}( \boldsymbol{x_{*}})$ and variance $\mathbb{V}(\boldsymbol{x}_{*})$ as:
\begin{equation}
    \Sigma = (k_{\text{RBF}}(\boldsymbol{X},\boldsymbol{X}) + \sigma^2\boldsymbol{I})^{-1}
\end{equation}
\begin{equation}
    \overline{g}( \boldsymbol{x_{*}}) = k_{\text{RBF}}(\boldsymbol{X}, \boldsymbol{x}_{*})^{\text{T}}\Sigma \boldsymbol{Y}
\end{equation}
\begin{equation}
    \mathbb{V}(\boldsymbol{x}_{*}) = k_{\text{RBF}}(\boldsymbol{x}_{*},x)^{\text{T}} \Sigma k_{\text{RBF}}(\boldsymbol{x}_{*}, x)^{\text{T}}
\end{equation}

\subsection{Tactile Servoing and Perception using Multi-fingered Hand \label{sliding_ada}}
Our one-leader-multi-follower multi-finger tactile sensing strategy mainly consists of two parts: (1) Hand Palm twist for tactile servoing, which is based on the tactile feature provided by the \textbf{``leader"} fingertip, and (2) \textbf{``follower"} fingertips adapt their poses to acquire larger sensing area based on their tactile features.

\textbf{Palm Twist for Tactile Servoing}: Based on the tactile feature $t_f = [ \overline{x}, \overline{y}, c]^T$ extracted from the ``leader" fingertip, the goal of this phase is to minimize the distance $\Delta t_f = [\Delta x, \Delta y, \Delta c]^T$ between current $t_f$ and reference $t_{fr} = [\frac{h}{2}, \frac{w}{2}, c_r]^T$ only by hand twisting. \textit{The joints of each finger will remain unchanged.} The hand twist motion $V_{palm} = [\dot{\theta_y}, \dot{\theta_z}, \dot{d_x}]^T$ is mapped from $\Delta t_f$ by a \textit{task-dependent} tactile Jacobian $J_s$ and \textit{task-dependent} selection matrix $P$ as introduced by \cite{li2013control}. In which $R_y$ and $R_z$ denote hand palm rotation angle along the Y axis and Z axis respectively and translation distance along the X axis w.r.t. the ``leader" fingertip frame
\begin{equation}
    J_s^{palm} = 
    \begin{bmatrix}
        \frac{\partial\overline{x}}{\partial \theta_z} & 0 & 0\\
        0 & \frac{\partial\overline{y}}{\partial \theta_y} & 0 \\
        0 & 0 & \frac{\partial c}{\partial d_x}
    \end{bmatrix} \quad P = \begin{bmatrix} i & 0 &0 \\ 0 & i & 0 \\ 0 & 0 & j      
    \end{bmatrix}.
\label{jac_selec}
\end{equation}
We set $i=0, j=1$ for translation mapping and $i=1, j=0$ for rotation mapping. The core mapping is formulated as 
\begin{equation}
    V_{palm} = PJ_s^\dagger\Delta t_f
\end{equation}
where $\dagger$ denotes the pseudo inverse. A PD controller is adopted for decreasing the $\Delta t_f$.

\textbf{Fingertip Adaptation for Tactile Perception}: Based on the tactile feature extracted from the ``follower" fingertips, the goal of this phase is to sense a larger contact area by adjusting the ``follower" fingertips' pose only by moving the corresponding finger joint position. 
\textit{The hand palm's pose will remain fixed at this moment.} Different from the above tactile servoing phase, due to the joint control accuracy and manipulability of the Allegro hand, this fingertip adaptation motion will directly change the fingertips' pose without feedback control and only be triggered if the fingertip is in contact. A similar task-dependent Jacobian and selection matrix as (\ref{jac_selec}) is pre-defined for fingertip pose adaptation. 

\subsection{Constrained Bayesian Optimization for Sliding Touch}
Bayesian Optimization (BOpt) is used to optimize black-box functions that are expensive to evaluate. In the context of this paper, the target (black-box) function is based on the implicit function that represents the object surface.
BOpt builds a (probabilistic) surrogate model, GPIS, of the target function and uses an acquisition function to determine where to evaluate next. We approximate the target function by optimizing our GPIS through sequential tactile perception.

\textbf{Surrogate Model}:
The surrogate model is a statistical model to describe the target function based on observations, and also provides predictive uncertainty. In our case, the GPIS is the surrogate model for the target function approximation and description. Given a query point $x_q$ from the world frame, our surrogate model provides a mean estimate $\bar{g}(x_q)$ and the corresponding variance $\mathbb{V}(x_q)$. The mean value defines whether or not the given query point is on the object's surface based on (\ref{gp_setting}) and the variance value denotes the prediction uncertainty.

\textbf{Acquisition Function}: 
An acquisition function is designed for balancing the exploration of new parameters vs the exploitation of current knowledge based on previous experiments. We choose Expected-Improvement (EI) as our acquisition function:
\begin{equation}
\text{EI}(x) = \mathbb{E}(\max(y - g_{best}, 0)), \quad y \sim g(x),
\end{equation}
where $g_{best}$ is the best-seen value of the black-box function. 
The next query point is computed by
\begin{equation}
    x_{n+1} = \arg \max_{x} \text{EI}_n(x).
\end{equation}
EI calculates how much its function value can be expected to improve over our current optimum for every point from our constrained search space. Then we will choose the point with maximal improvement as our query point for the next sliding touch guidance. 

\textbf{Constrained Search Space}: 
To improve the efficiency of sampling query points and well integrate with the hand sliding motion, the search space is updated with the GPIS for each iteration. The search space $S^x$ for each sampling is constrained by the implicit surface
\begin{equation}
    S^x_{n+1} = \{ x  \mid \arg g^{\text{IS}}_n(x)=0 \wedge \lVert x - x^{ob}_n\rVert > d \}
\end{equation}
where $d=0.005$ denotes the minimized length between points from search space and observed points.

\subsection{Surface Exploration and Exploitation}

BOpt is designed to output a query point $x_{BO} \in \mathbb{R}^3$ with respect to the world frame in each iteration. Combined with the current ``leader" fingertip's tactile feature, the output $x_{BO}$ will guide the robotic hand's sliding servoing direction to explore the target object's surface. Given the estimated coordinate of the center of intensity $x_c=[i,j,v]^T \in \mathbb{R}^3$ and its corresponding surface normal $n \in \mathbb{R}^3$ from the ``leader" fingertip w.r.t. the world frame, the sliding direction $t_{BO}$ is the projection of $x_{BO}-x_c$ on the estimated surface (defined by normal n), which is calculated as 
\begin{equation}
t_{BO} = (x_{BO}-x_c) - \frac{(x_{BO}-x_c) \times n}{n\times n}  n.
\end{equation}
The translation for the next frame is computed as 
\begin{equation}
    x_{next} = x_c + \frac{t_{BO}}{\|t_{BO} \|}\Delta step,
\end{equation}
where the $\Delta step = 0.005$ is the step size between each sliding frame and needs to be set small enough to keep the ``leader" fingertip in contact with the target object.

\begin{figure} [htb]
    \centering
    \includegraphics[width=\linewidth]{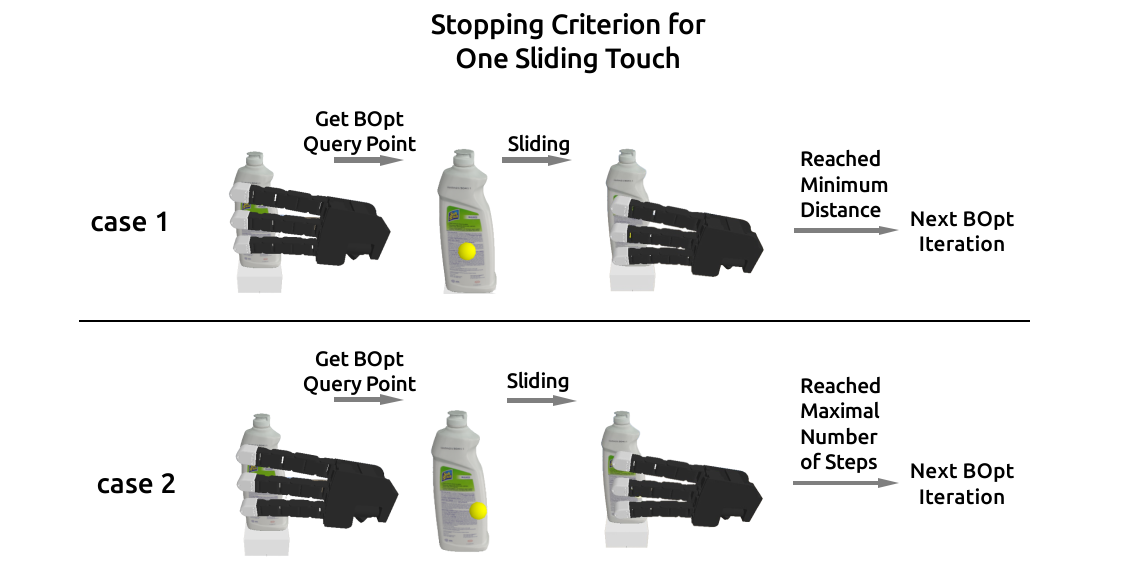}
    \caption{Two different examples of how stopping criterion is triggered.}
    \label{bo_stop}
\end{figure}

\textbf{Sliding Exploration for each BOpt iteration}:
One sliding touch consists of sequential touch frames as introduced in (\ref{formualtion}). Each frame is obtained from every single step, and the total number of frames for the ongoing sliding touch is determined by (1) the distance between the current center of contact intensity from the ``leader" fingertip and the query point smaller than the minimal value $D_{min} = 0.01$, (2) the number of frames reached the maximal boundary $N_{max} = 15$. If one of the two conditions is met, the ongoing sliding touch will terminate and enter the next BOpt iteration as shown in Fig. \ref{bo_stop}.

\textbf{Stopping Criterion for Surface Exploration}:
The goal of surface exploration is to estimate the target object's shape with relatively low global uncertainty. This is reflected in the uncertainty distance reduction ratio (UDRR) of our GPIS model.
{\color{black}{Since the GPIS yields low uncertainty in the areas with dense observation points, e.g. points obtained from the vision sensor, the global estimation is formed based on comparison with the areas that are partially observed. As the gap between the global highest uncertainty and the uncertainty of visually perceived areas decreases, we consider that the area we are exploring becomes increasingly confident. Given the initial GPIS learned from only the visual point cloud, the initial uncertainty distance is calculated as:}}
\begin{equation}
\begin{aligned}
    dist_{max} = \max(\mathbb{V}_{init}(x_{m}) - \mathbb{V}_{init}(x_{n})),  \quad x_{m}, x_{n} \in \mathbf{P^w_d}
\end{aligned}
\end{equation}
After the sliding touch of each BOpt iteration is over, the outcome UDRR is calculated as follows
\begin{equation}
    \text{UDRR} = \frac{\max(\mathbb{V}(x_{m}) - \mathbb{V}(x_{n}))}{dist_{max}},  \quad x_{m}, x_{n} \in \mathbf{P^w}
\end{equation}
If $\text{UDRR} < 30\%$, we consider the target object has been explored sufficiently and the sliding perception is over. However, a lower UDRR threshold would lead to a more confident shape modeling.

\section{Experimental Validations \label{sec_6}}
We evaluate our proposed sliding strategy in simulation (Section \ref{sim_exp}) and the whole pipeline in the real world (Section \ref{real_exp}). We set up our robotic platform with a Kuka iiwa 7 with Allegro Hand as shown in Fig. \ref{allegrohand}. An Azure Kinect DK depth camera is used as a scene camera for visual perception. All test objects are from the YCB \cite{calli2015ycb} and the OCRTOC benchmark datasets \cite{liu2021ocrtoc}. The shape estimation performance is quantified using a shape similarity metric, the Chamfer distance (CD) \cite{barrow1977parametric}.

\textbf{Implementation}: The whole pipeline is executed on an Intel Core i7-11800H CPU, 16 RAM with an NVIDIA GeForce RTX 3070 GPU. To balance the point cloud density from visual and tactile perception, the tactile point cloud is randomly down-sampled with a ratio of 0.02. Finally the input point cloud is down-sampled to 4500 points. During the GPIS evaluation phase, we set the grid size with a value of 10 for surface reconstruction. 

\begin{figure} [h]
   \centering
\includegraphics[width=0.8\linewidth]{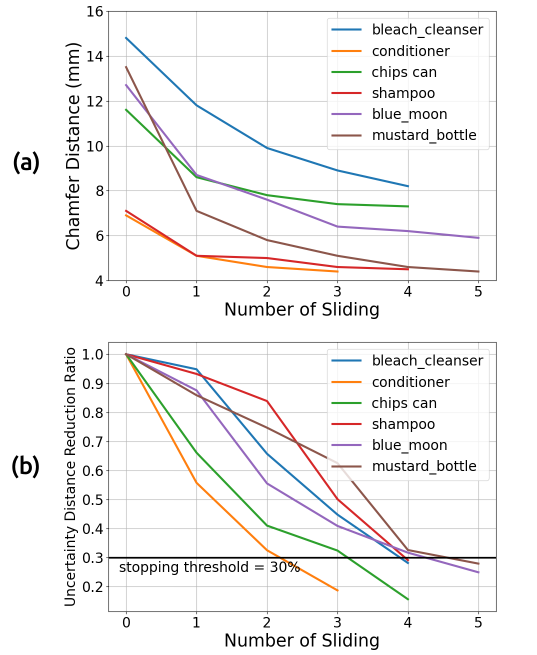}
    \caption{The figure illustrates shape estimation accuracy (a) and the UDRR (b) of 6 objects in simulation with respect to the number of sliding touches.}
    \label{sim_cd}
\end{figure}

\subsection{Sliding Simulation \label{sim_exp}}
We tested our sliding touch strategy in simulation with 6 different objects, which are of varying shapes, curvature, and sizes to verify the generalization of the proposed method. The performance of our method is evaluated by (1) the accuracy of estimated meshes and (2) the efficiency of object exploration. The sliding touch exploration is visualized in Fig. \ref{sim_results}, which shows the tactile point cloud obtained from every single sliding touch colored differently for clarity in the first column. The evolution of the global uncertainty and estimated mesh during the perception process can be seen in the middle columns. For objects with non-symmetric shapes, we test our method under two different camera views (the rotation angle between these 2 views is larger than $60^{\circ}$) for a comprehensive evaluation. Our method output the estimated shape with considerably reduced uncertainty, with respect to the initial model built based on only visual data, only using a few continuous touches. The change in the uncertainty map clearly demonstrates that each sliding touch increases the information gain and helps to efficiently capture the surface's underlying geometry. The accuracy of the estimated shape calculated using CD and the evolution of UDRR is shown in Fig. \ref{sim_cd} (a) and (b), respectively. As the number of sliding touches increases, CD w.r.t the ground truth model decreases for each object. For those objects with irregular or non-symmetric shapes in particular, e.g. mustard bottle and bleach cleanser, in which GPIS has a relatively bad initial estimation due to partial and vision-only observation points, it is seen from a large amount of decrease in CD that our proposed method can efficiently explore the areas (not visible to the camera) and provides sufficient estimated shape correction.

We compare the efficiency of our method with the constrained random policy. To fairly evaluate both policies, \textit{the search space for the constrained random policy is set the same as our BOpt}. The exploration is ended by the same stopping criterion with a value of UDRR lower than $30\%$. The comparison result is listed in Table \ref{table_1}. Even though both policies under the constrained search space can perform full exploration on the target object's region, our proposed BOpt approach achieves similar accuracy with a significantly lower number of sliding touches. The smaller standard deviation of BOpt iteration times also demonstrates that our method is more stable compared to the baseline.

\begin{table}[h]
\footnotesize
\scriptsize
\caption{Performance Comparison Between BOpt and Random Policy}
\begin{tabular}{|l|ll|ll|}
\hline
Objects &       \quad Avg., Std.& Num. $^1$              &  \quad Avg., Std.  &  CD$^2$ \\ \cline{2-5} 
 ~ & \multicolumn{1}{l|}{BOpt (Ours)} & Random$^*$ & \multicolumn{1}{l|}{BOpt (Ours)} & Random$^*$ \\ \hline
 $\text{Shampoo}^\text{b}$ & \multicolumn{1}{l|}{\textbf{3.3, 0.48}} & 4.4, 1.43 & \multicolumn{1}{l|}{4.33, \textbf{0.11}} & 4.29, 0.14 \\ \hline
 $\text{Chips Can}^\text{a}$ & \multicolumn{1}{l|}{\textbf{4.5, 0.67}} & 6.2, 1.54 & \multicolumn{1}{l|}{\textbf{7.52}, 0.23} & 7.57, 0.21 \\ \hline
 $\text{Bleach Cleanser}^\text{a}$& \multicolumn{1}{l|}{\textbf{4.7, 0.64}} & 5.7, 1.10 & \multicolumn{1}{l|}{8.25, 0.31} & 8.21, 0.24\\ \hline
$\text{Conditioner}^\text{b}$ & \multicolumn{1}{l|}{\textbf{3.2, 0.40}} & 5.3, 1.50 & \multicolumn{1}{l|}{4.45, \textbf{0.17}} & 4.34, 0.20 \\ \hline
$\text{Blue Moon}^\text{b}$ & \multicolumn{1}{l|}{\textbf{5.0, 0.45}} & 7.2, 0.98 & \multicolumn{1}{l|}{6.03, 0.34} & 5.98, 0.27 \\ \hline
 $\text{Mustard Bottle}^\text{a}$& \multicolumn{1}{l|}{\textbf{4.8, 0.75}} & 6.7, 0.90 & \multicolumn{1}{l|}{\textbf{4.72, 0.45}} & 4.77, 0.51 \\ \hline
\end{tabular}
\begin{tablenotes} 
		\item $^1$ Number of sliding touch times to achieve $\text{UDRR} < 30\%$  from 10 experiments.
  \item $^2$ Output shape's Chamfer Distance (mm) w.r.t GT from 10 experiments.
  \item $^\text{a}$ and $^\text{b}$ respectively denotes from the YCB and OCRTOC benchmark.
  \item $^*$ The constrained random policy shares the same constrained search space.
\end{tablenotes} 
\label{table_1}
\end{table}

\subsection{Real-world Experiment \label{real_exp}}
\begin{figure}[h]
    \centering
    \includegraphics[width=\linewidth]{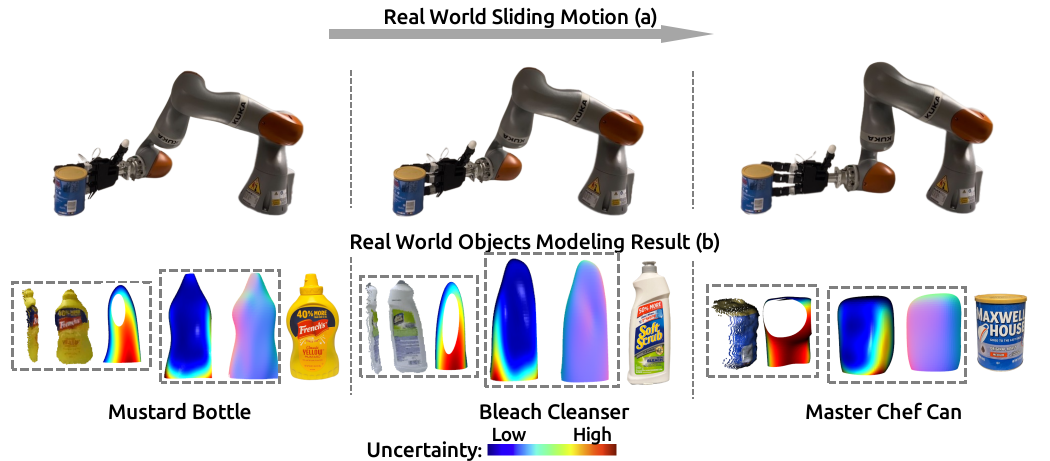}
    \caption{Figures from the top line show the sliding motion of one complete experiment on Master Chef Can. Figures from the bottom line show the visual-only point cloud with the initially estimated shape, the final estimation result with an uncertainty map, and the corresponding real-world object separately for each object. For objects from left to right, the exploration process takes respectively 5, 4, and 4 sliding touches to produce the resulting output mesh.}
    \label{real_world_result}
\end{figure}

We tested our proposed method with real-world objects from the YCB dataset. We selected three objects with different materials, including symmetric and non-symmetric shapes with different sizes. The initial approach position is set on the left side of the target object w.r.t camera frame, and the example of one complete sliding sensing is shown in Fig. \ref{real_world_result} (a). Fig. \ref{real_world_result} (b) shows the qualitative result of the shape reconstruction on three objects. Due to safety concerns for avoiding collision with the table, the coordinate of the query point will be limited with a Z value between 0.05 and 0.15 w.r.t the world frame. The proposed method yields reconstructed shapes for unknown objects with low uncertainty and detailed curvature, capturing the main geometric details in explored areas. However, there are still some areas that are not well explored due to arm reachability, which are left with relatively high uncertainty.
\section{CONCLUSIONS \label{sec_7}}
We present a novel approach for unknown object modeling based on multi-fingered hand sliding touches. To efficiently address this problem, an essential aspect is to fully utilize the local shape sensing ability of each fingertip and integrate them in a continuous and smooth manner. Our method yields successful reconstruction results capturing the underlying shape in visually unseen areas with only a few continuous sliding touches. However, our work relies on several assumptions due to the limitation of the real setup. First, the objects are fixed on the table to be able to apply a larger contact force to generate sufficient deformation on the gel of tactile sensors. Second, the explored region on the object's surface is considered low curvature to meet the limited reachability of the robotic arm. In order to address these issues, we plan to combine the proposed approach with a mobile manipulator, which could provide a larger workspace, and sensors with higher sensitivity and precision.




\bibliographystyle{IEEEtran}
\bibliography{ref}
\end{document}